\documentclass[letterpaper, 10 pt, journal, twoside]{IEEEtran}
\ifCLASSINFOpdf
\else
\fi
\usepackage{times}
\usepackage{epsfig}

\usepackage{color} 
\usepackage{amsmath}
\usepackage{amssymb}
\usepackage{graphicx}
\usepackage{multirow}
\usepackage{tabularx}
\usepackage{adjustbox}
\ifCLASSOPTIONcompsoc
 \usepackage[caption=false,font=normalsize,labelfont=sf,textfont=sf]{subfig}
\else
 \usepackage[caption=false,font=footnotesize]{subfig}
\fi
\usepackage{siunitx}
\usepackage{textcomp}

\usepackage{stmaryrd} 
\usepackage{latexsym} 

\usepackage{url}
\usepackage{hyperref} \hypersetup{colorlinks=true, linkcolor=black, urlcolor=RubineRed}
\usepackage{kotex} 

\newcommand{\figref}[1]{Fig.~\ref{#1}}

\newcommand{\tabref}[1]{Tab.~\ref{#1}}
\newcommand{\eqnref}[1]{Eq.~(\ref{#1})}

\newcommand{\ie}{\textit{i.e.}}
\newcommand{\eg}{\textit{e.g.}}
\newcommand{\etal}{\textit{et al.}}

\usepackage[dvipsnames]{xcolor}

\begin{document}
%
\title{Self-supervised Depth and Ego-motion Estimation for Monocular Thermal Video using \\Multi-spectral Consistency Loss}
%
%
%


\author{Ukcheol Shin, Kyunghyun Lee, Seokju Lee, and In So Kweon %
\thanks{Manuscript received: September 5, 2021; Revised December, 8, 2021; Accepted January, 10, 2022.
This letter was recommended for publication by Associate Editor R.Ranftl and Editor C.Cadena Lerma upon evaluation of the
reviewers’ comments. 
This work was supported by the International Research, and Development Program of the National Research Foundation of Korea funded by the Ministry of Science and ICT under Grant NRF-2021K1A3A1A21040016.
\textit{(Corresponding author: In So Kweon.)}}
\thanks{U. Shin, K. Lee, S. Lee, and I. S. Kweon are with the School of Electrical Engineering, KAIST, Daejeon 34141, Republic of Korea.
{\tt \{shinwc159, kyunghyun.lee, seokju91, iskweon77\}@kaist.ac.kr}}%
\thanks{Digital Object Identifier (DOI): see top of this page.}
}
%
%

\markboth{IEEE Robotics and Automation Letters. Preprint Version. Accepted December, 2021}
{Shin \MakeLowercase{\textit{et al.}}: Self-supervised Depth and Ego-motion Estimation for Monocular Thermal Video using Multi-spectral Consistency} 

%



\maketitle
\begin{abstract}
A thermal camera can robustly capture thermal radiation images under harsh light conditions such as night scenes, tunnels, and disaster scenarios. 
However, despite this advantage, neither depth nor ego-motion estimation research for the thermal camera have not been actively explored so far.
In this paper, we propose a self-supervised learning method for depth and ego-motion estimation from thermal images.
The proposed method exploits multi-spectral consistency that consists of temperature and photometric consistency loss.
The temperature consistency loss provides a fundamental self-supervisory signal by reconstructing clipped and colorized thermal images. 
Additionally, we design a differentiable forward warping module that can transform the coordinate system of the estimated depth map and relative pose from thermal camera to visible camera.
Based on the proposed module, the photometric consistency loss can provide complementary self-supervision to networks.
Networks trained with the proposed method robustly estimate the depth and pose from monocular thermal video under low-light and even zero-light conditions.
To the best of our knowledge, this is the first work to simultaneously estimate both depth and ego-motion from monocular thermal video in a self-supervised manner. 
\end{abstract}


%
\IEEEpeerreviewmaketitle

%
%
%
%
\section{Introduction}

\IEEEPARstart{S}{elf-supervised} learning of depth and ego-motion estimation is an actively researched topic to train a neural network without relying on ground-truth depth and pose labels, which require expensive equipment (\eg, Lidar and motion capture system) and a complicated label generation process.
However, most self-supervised depth and ego-motion networks~\cite{zhou2017unsupervised,yin2018geonet,godard2019digging,bian2019unsupervised} have been designed for visible cameras.
Their performance is not guaranteed under low-light conditions such as dark rooms, tunnels, and night road scenes. 
This is because visible cameras typically introduce noise, motion blur, and undesirable exposure levels in these low-light scenarios~\cite{shin2019camera}.
Furthermore, illumination condition in the real world unexpectedly change depending on the weather, time, and location. 
Therefore, visible image based networks are difficult to use in the real-world due to their inherent sensitivity to lighting conditions.

\begin{figure}[t]
\begin{center}
\footnotesize
\begin{tabular}{c@{\hskip 0.005\linewidth}c@{\hskip 0.005\linewidth}c@{\hskip 0.005\linewidth}c@{\hskip 0.005\linewidth}}
\multicolumn{4}{c}{\includegraphics[width=0.96\linewidth]{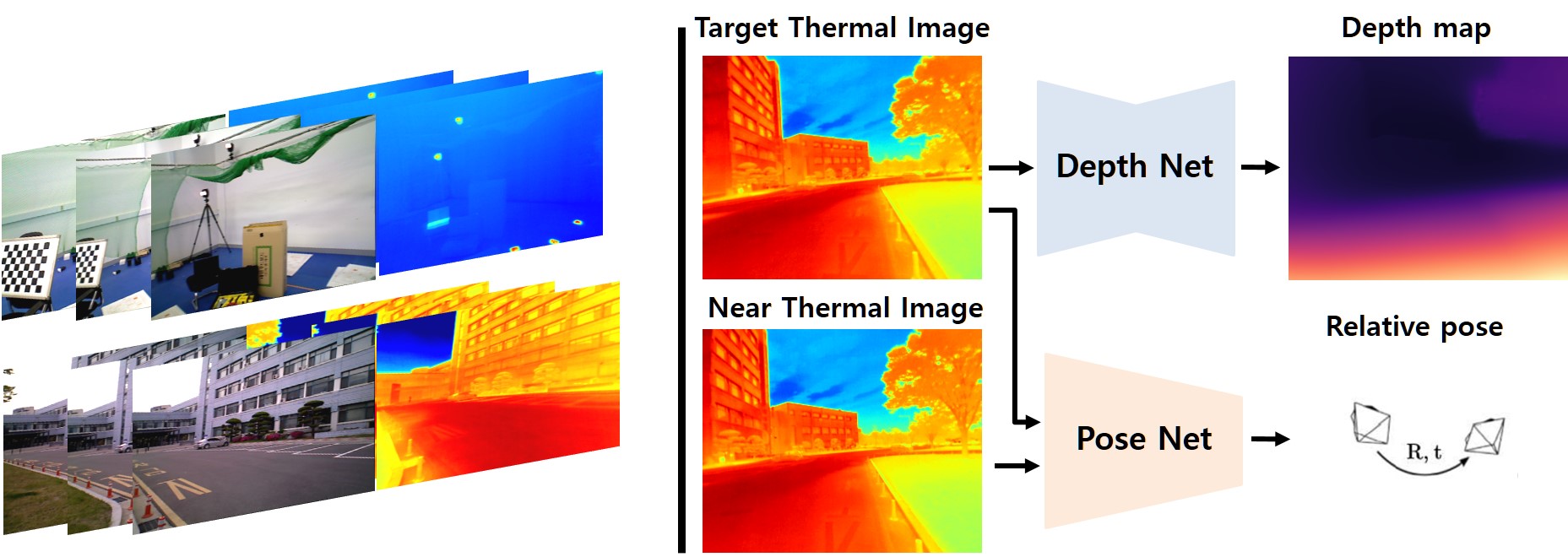}} \\ 
\multicolumn{2}{c}{\bf (a) Training: } & \multicolumn{2}{c}{\bf (b) Testing: } \\ 
\multicolumn{2}{c}{\bf Unlabeled RGB-T video.} & \multicolumn{2}{c}{\bf Monocular thermal video.} 
\vspace{0.05in} \\
\includegraphics[width=0.24\linewidth]{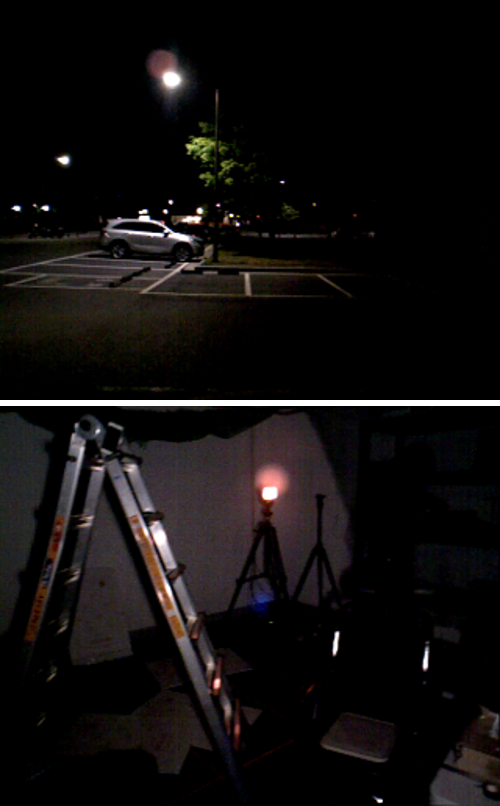} &  
\includegraphics[width=0.24\linewidth]{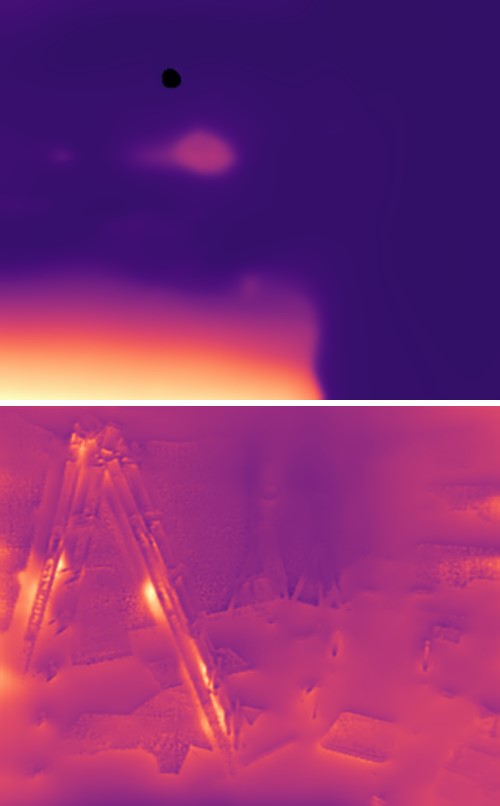} &  
\includegraphics[width=0.24\linewidth]{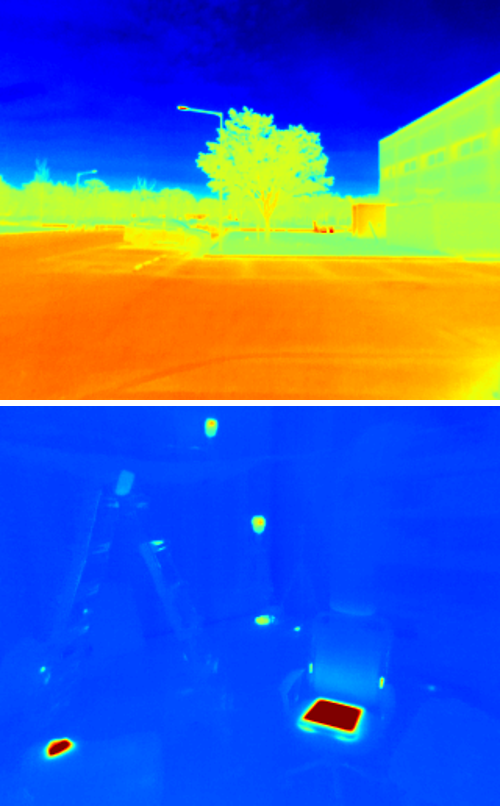} &  
\includegraphics[width=0.24\linewidth]{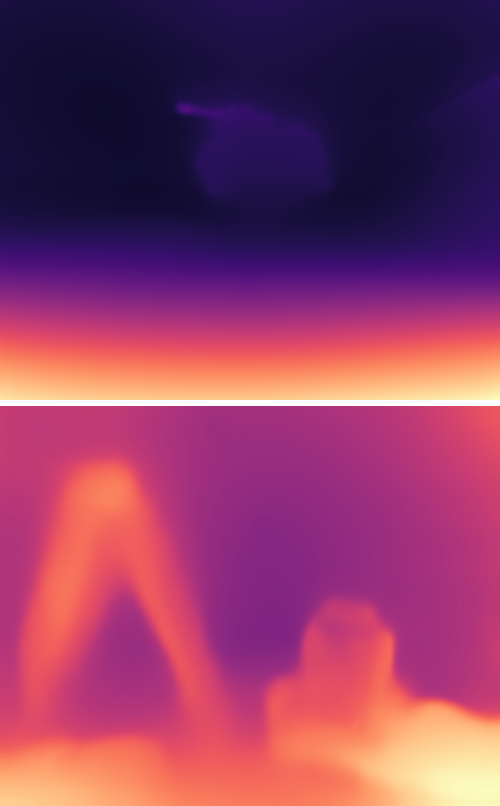} \\
{\bf RGB image} & {\bf Bian~\etal~\cite{bian2019unsupervised}.} & {\bf Thermal image} & {\bf Ours}  \\
\end{tabular}
\end{center}
\caption{{\bf Overview of proposed self-supervised depth and pose learning methods for thermal images}. 
In the training step (a), the networks are trained with an unlabeled visible-thermal video. 
In the testing step (b), single-view depth and pose are estimated from monocular thermal image sequences. 
The proposed networks robustly estimate the depth and pose under low- and zero-light conditions.
}
\label{fig:teaser}
\end{figure}

A thermal camera is one potential solution thanks to its environment insensitive property; it can directly measure long-wave infrared radiation of objects regardless of the presence of an external light source. 
Therefore, the thermal camera can capture consistent image data under various light and weather conditions. 
However, despite this advantage, a few unique characteristics of thermal images hinder their effectiveness.
Unlike visible images, thermal images have relatively low resolution, low signal-to-noise ratio, low contrast, and blurry edges. 
These properties weaken a self-supervisory signal from the image reconstruction mechanism commonly used in self-supervised learning methods.

Furthermore, proper thermal image format for temporal thermal image reconstruction has been not explored so far.
Various recognition tasks~\cite{treible2017cats,choi2018kaist,sun2020fuseseg,kim2021ms} utilize 8bit thermal image format that have already lost temporal consistency between adjacent images through the built-in signal processing pipeline of the thermal camera.
On the other hand, 14bit RAW sensor data preserve temporal consistency.
However, these data can not be directly utilized for image reconstruction loss due to the wide measurement range.
For example, the RAW sensor data can represent temperatures in an approximately -30\textdegree C to 150 \textdegree C range in low-gain mode~\cite{flir-ax5}. 
On the other hand, the temperature distribution in the real-world has a small variance about of ${\pm}$10\textdegree C.

\begin{figure}[t]
\begin{center}
    \footnotesize
    \begin{tabular}{c@{\hskip 0.004\linewidth}c@{\hskip 0.004\linewidth}c@{\hskip 0.004\linewidth}c@{\hskip 0.004\linewidth}}
     \includegraphics[width=0.24\linewidth]{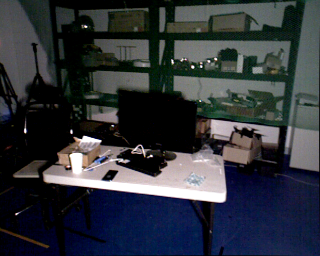} &
     \includegraphics[width=0.24\linewidth]{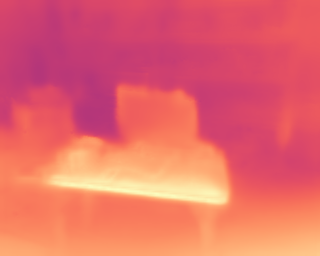} &
     \includegraphics[width=0.24\linewidth]{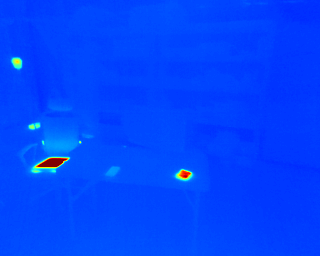} &
     \includegraphics[width=0.24\linewidth]{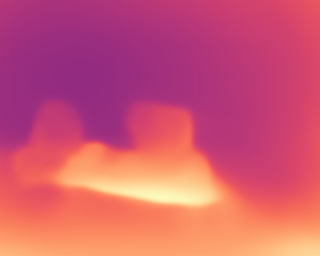} \vspace{-0.03in}\\
     \includegraphics[width=0.24\linewidth]{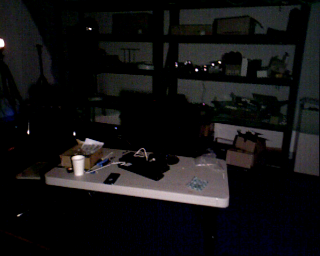} &
     \includegraphics[width=0.24\linewidth]{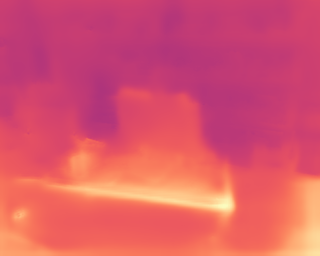} &
     \includegraphics[width=0.24\linewidth]{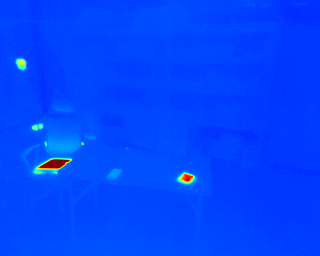} &
     \includegraphics[width=0.24\linewidth]{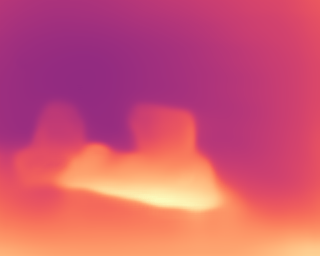} \vspace{-0.03in}\\
     \includegraphics[width=0.24\linewidth]{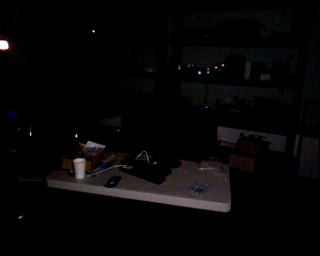} &
     \includegraphics[width=0.24\linewidth]{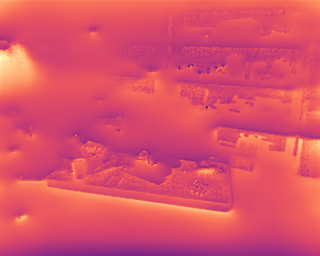} &
     \includegraphics[width=0.24\linewidth]{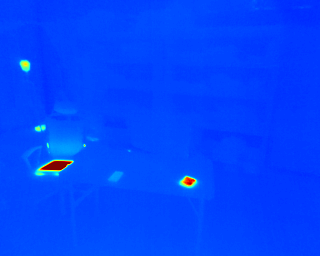} &
     \includegraphics[width=0.24\linewidth]{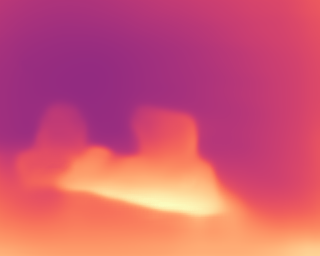} \vspace{-0.03in} \\
     {\footnotesize(a) RGB} & {\footnotesize (b) Depth(RGB) } & {\footnotesize (c) Thermal} & {\footnotesize (d) Depth(ours)} \\
    \end{tabular}
\end{center}
\caption{{\bf Depth estimation results from RGB and thermal images}.  Depth estimation from RGB images easily degrades according to light-condition. However, depth estimation from thermal images shows robust prediction results regardless of light-condition change. 
}
\label{fig:rgb_thermal_comparison}
\end{figure}

In this paper, we propose a self-supervised learning method that can jointly train single-view depth and multi-view ego-motion networks from visible-thermal video, as shown in ~\figref{fig:teaser}.
After training, only the thermal image sequence is utilized in the testing step to estimate depth map and pose.
The proposed method effectively handles the problems mentioned above by exploiting multi-spectral consistency loss.
Our contributions include the following:
\begin{itemize}
\item
We propose an self-supervised learning method that exploits the temperature and photometric consistency loss for single-view depth and multi-view pose estimation from a monocular thermal video.
\item 
We propose an efficient thermal image representation strategy, called clipping-and-colorization, that can provide a sufficient self-supervisory signal for temperature consistency loss while preserving temporal consistency.
\item
We design a differentiable forward warping module that can transform the depth map and pose from the thermal image coordinate system to the visible image coordinate system. 
Based on the proposed module, the photometric consistency loss can synthesize a visible image with depth map and pose estimated from thermal images.
\item
We demonstrate that networks trained with the proposed method robustly and reliably estimate depth and pose results under low-light and even zero-light conditions, as shown in~\figref{fig:rgb_thermal_comparison}.
\end{itemize}

\section{Related Works}
\label{sec:related works}
{\bf Self-supervised Depth and Ego-motion Networks.} 
With the emergences of the deep neural network, lots of learning-based depth and ego-motion networks have been proposed ~\cite{zhou2017unsupervised,zhou2018deeptam,yin2018geonet,godard2019digging,ranftl2020towards,lee2021attentive}. 
SfM-learner~\cite{zhou2017unsupervised} is a pioneering work that demonstrates that depth and pose estimation networks can be trained in a fully unsupervised manner from monocular video. 
However, their performance is limited because they assume a static scene, brightness consistency, and Lambertian surface for image reconstruction. 
Also, the inherent scale-ambiguity and scale-drift problem of monocular video hinder prediction of long-term camera trajectory.
To exclude dynamic objects, SfM-learner~\cite{zhou2017unsupervised} utilizes an explainability mask and object mask.
GeoNet~\cite{yin2018geonet} and Ranjan~\etal~\cite{ranjan2019competitive} explicitly separate rigid flow and object motion flow using an additional optical flow network or motion segment network. 
For the scale-ambiguity problem, several works~\cite{mahjourian2018unsupervised,bian2019unsupervised,chen2019self} impose geometric constraints to predict a scale-consistent depth and camera trajectory.

In contrast to the research mentioned above, which focuses on the RGB domain, our proposed methods focus on self-supervised learning of depth and pose estimation for the thermal domain, which have not been actively explored so far. 
However, we found that the scale inconsistent depth and invalid image reconstruction sources, such as moving objects and static camera motion, also existed in the thermal domain. 
Therefore, we exploited geometric constraints and excluded invalid pixels to handle these problems.

{\bf Depth and Ego-motion Estimation with Thermal Image.}
A thermal camera can robustly capture temperature information regardless of lighting and weather conditions.
However, thermal images generally suffer from inherent weak image properties such as low contrast, less texture, and low resolution.
Therefore, most traditional depth or pose estimation studies~\cite{khattak2020keyframe, shin2019sparse,delaune2019thermal,borges2016practical} utilize other heterogeneous sensors as well, such as IMU, visible cameras, and Lidar, to complement these properties.
Also, a few existing neural network based depth and pose estimation studies have also utilized RGB sources~\cite{kim2018multispectral} or direct pose supervision~\cite{saputra2020deeptio}.
Kim~\etal~\cite{kim2018multispectral} proposed an unsupervised multi-task learning framework that utilizes chromaticity clues and RGB-based photometric error to estimate a depth map from a thermal image. 
However, for this purpose, they carefully designed an RGB stereo and thermal camera system in which the principal points of the thermal camera and one RGB camera are geometrically aligned utilizing an XYZ stage and beam-splitter.
DeepTIO~\cite{saputra2020deeptio} proposed a thermal-inertial odometry network trained in a supervised manner.
They trained a thermal image's feature encoding network to mimic the RGB image's encoding network.

Among the above-mentioned studies, the work most related to ours is Kim~\etal~\cite{kim2018multispectral}.
However, our method differs from their method in three aspects.
First, their method utilizes the spatial relation between left-right RGB images, while our method exploits temporal relation.
Second, their method does not utilize thermal image reconstruction loss.
Lastly, their method requires a geometrically aligned RGB-thermal camera system, which is infeasible in most cases.
On the other hand, our methods can utilize any RGB-Thermal camera system with an arbitrary geometric relationship.
 
\begin{figure*}[t]
\begin{center}
\footnotesize
\begin{tabular}
{c@{\hskip 0.02\linewidth}c}
\includegraphics[width=0.60\linewidth]{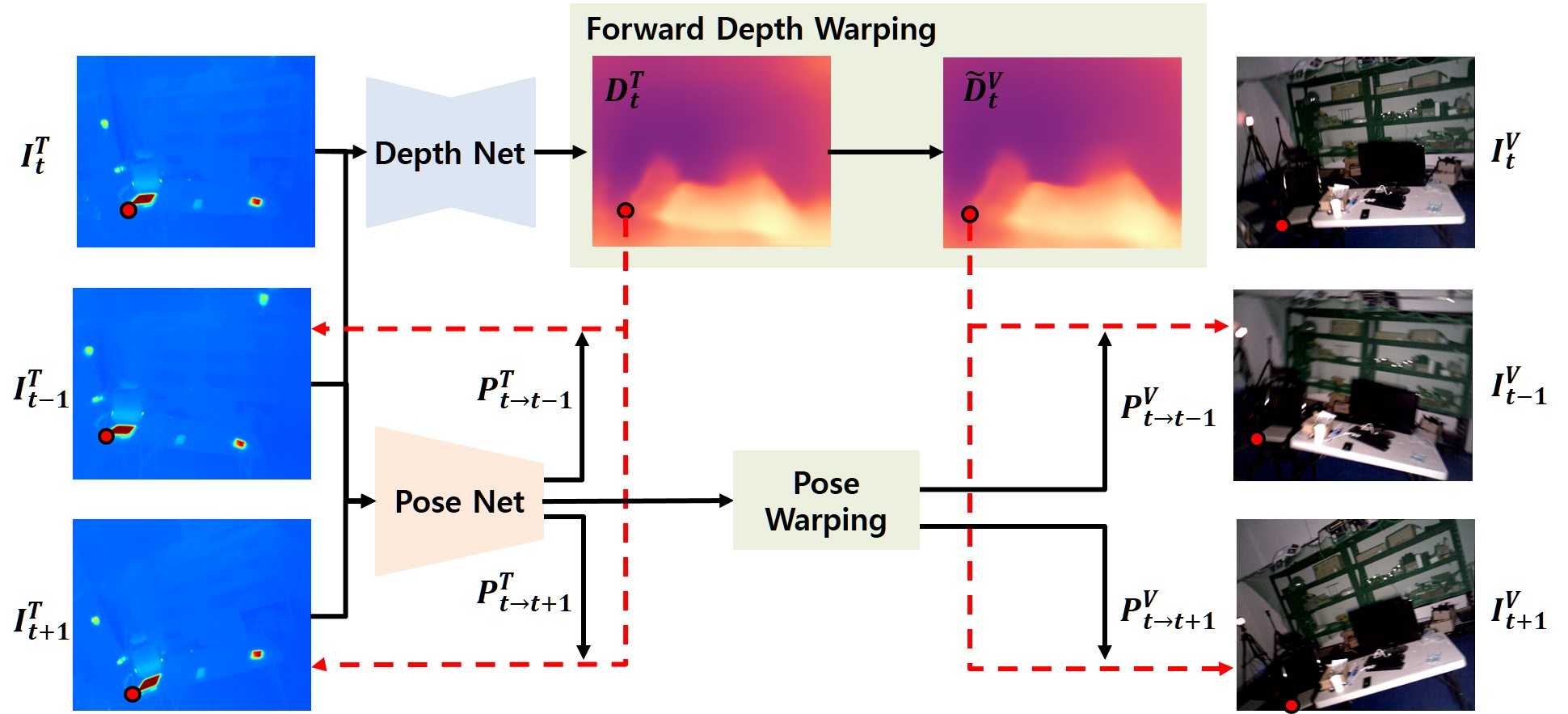} &
\includegraphics[width=0.38\linewidth]{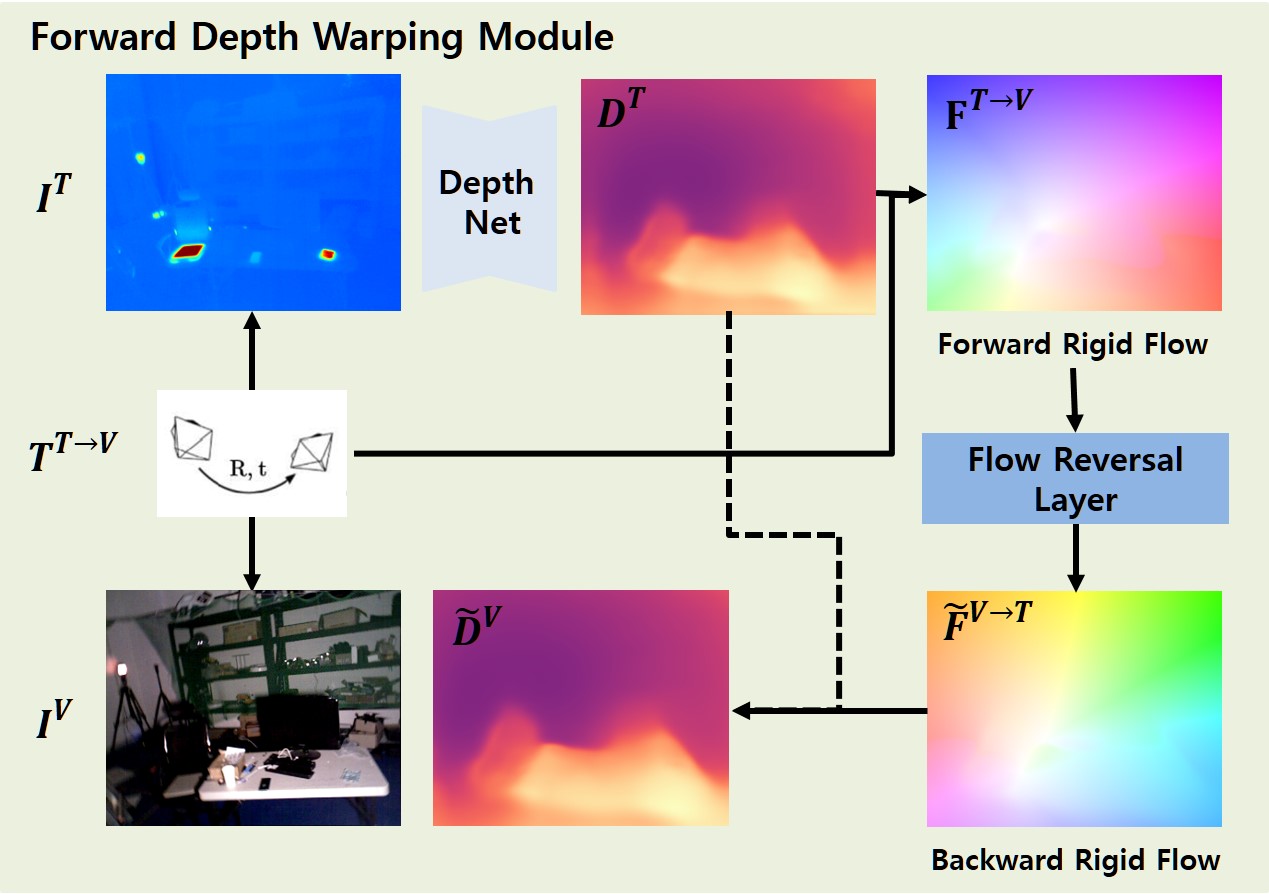} \\
{\footnotesize (a) Pipeline of the multi-spectral consistency loss} & {\footnotesize (b) Pipeline of the forward depth warping module} \\
\end{tabular}
\end{center}
\caption{{\bf Overall pipeline of proposed multi-spectral consistency loss}. 
Multi-spectral consistency loss consists of temperature consistency loss and photometric consistency loss. 
The depth map $D_t^{T}$ and poses $P_{t \shortrightarrow{} t+1}^{T}$ are estimated from thermal images through the depth and pose network.
Thermal image is reconstructed with the depth map $D_t^{T}$ and pose $P_{t \shortrightarrow{} t+1}^{T}$. 
The depth map $\Tilde{D}_t^{V}$ and pose $P_{t \shortrightarrow{} t+1}^{V}$ of a visible image are generated from the forward depth warping and pose warping modules. 
Visible images are reconstructed with the depth map $\Tilde{D}_t^{V}$ and poses $P_{t \shortrightarrow{} t+1}^{V}$.} 
\label{fig:overall_net}
\end{figure*}

\section{Proposed Method}
\label{sec:method}

\subsection{Method Overview}
\label{sec:method_overview}
Our main objective is to train neural networks that can estimate accurate depth and pose from monocular thermal image sequences. For this purpose, we propose multi-spectral consistency loss to generate a self-supervision signal for the depth and pose networks, as shown in ~\figref{fig:overall_net}.
Overall objective function consists of a reconstruction loss $L_{rec}$ that minimizes synthesized image differences and the geometry consistency loss $L_{gc}$ that enforces the consecutive images have consistent 3D scene structures. 
These loss functions can be applied to both thermal and visual images. 
Therefore, our overall objective function can be formulated as follows: 
\begin{equation} 
\label{equ:total loss}
\begin{split}
L_{total}&=\lambda_{T}(\alpha L_{rec}^{T}+\beta L_{gc}^{T})+\lambda_{V}(\alpha L_{rec}^{V}+\beta L_{gc}^{V}),\end{split}
\end{equation}
where $\alpha$ and $\beta$ are hyper parameters. 
We utilize three sequential image pairs and both forward and backward direction loss in the training steps to maximize the data usage.
In the following sections, however, we use two consecutive image pairs $[(I_{t}^{T}, I_{t}^{V}),(I_{t+1}^{T}, I_{t+1}^{V})]$ for the simplified explanation.

\subsection{Temperature Consistency Loss}
{\bf Clipping-and-Colorization.}
The standard thermal camera can support two types of thermal images; 8bit and 14bit images.
However, the 8bit thermal image is a re-scaled image with the min-max value of the 14bit raw image through the thermal camera's image processing pipeline.
Therefore, the 8bit image loses the temporal consistency between adjacent images and is not suitable for self-supervised learning of depth and pose estimation; if two images have different min and max temperature values, the same object in two adjacent 8bit images has a different value.

Instead, we utilize a 14bit raw thermal image\footnote{Throughout the paper, we used the terminology "thermal image" as the 14bit raw radiometric image, which is convertible to temperature values.} to guarantee the temporal temperature consistency.
However, another problems are a wide measurement range and the aforementioned thermal image properties that weaken temporal image differences.
Therefore, we propose a simple but efficient thermal image normalization strategy, named clipping-and-colorization ($N_{CC})$.
The proposed strategy can be formulated as follows : 
\begin{equation} 
\label{equ:clliping-and-colorize Strategy}
I^{T,cc} = f_{color}\left(\frac{clip(I^{T,raw}, \tau_{min}, \tau_{max}) - \tau_{min}}{\tau_{max} - \tau_{min}}\right),
\end{equation}
where the function $clip(\cdot)$ clips the value of the raw thermal image in the pre-defined range ($\tau_{min}$, $\tau_{max}$).
The values ($\tau_{min}$ and $\tau_{max}$) are determined by the valid ranges of the training dataset and fixed during training.
The color mapping function $f_{color}(\cdot)$ converts the single-channel clipped image ($\mathbb{R}^{1\text{x}H\text{x}W}$) into a three-channel colorized image ($\mathbb{R}^{3\text{x}H\text{x}W}$) with the \textit{jet} color map table~\cite{Hunter:2007}. 

The proposed strategy can provide sufficient supervisory signals to the networks and achieve better edge-aware depth estimation results by enhancing the contrast of thermal images while preserving temporal consistency.

{\bf Thermal Image Reconstruction.}
Based on the strategy $N_{CC}$, we propose a temperature consistency loss to generate a self-supervisory signal for the depth and pose networks. 
Initially, the depth and pose networks estimate depth maps $(D_{t}^{T}, D_{t+1}^{T})$ and 6D relative camera pose $P_{t \shortrightarrow{} t+1}^{T}$ from consecutive thermal images $(I_{t}^{T}, I_{t+1}^{T})$. 
After that, a synthesized thermal image $\Tilde{I}_{t}^{T}$ is reconstructed with the source image $I_{t+1}^{T}$, the depth $D_{t}^{T}$, and the pose $P_{t \shortrightarrow{} t+1}^{T}$, in an inverse-warping manner~\cite{zhou2017unsupervised}. 
Based on the synthesized and original images, the image reconstruction loss can be formulated as follows : 
\begin{equation} 
\label{equ:recon loss}
\begin{split}
L_{rec}^{T}(p) = \gamma^{T} \frac{1-SSIM(I_{t}^{T,cc}(p), \Tilde{I}_{t}^{T,cc}(p))}{2} \\ + (1-\gamma^{T}) L_{1}(I_{t}^{T,cc}(p) ,\Tilde{I}_{t}^{T,cc}(p)),
\end{split}
\end{equation}
where $p$ denotes pixel coordinates, and $\gamma$ indicates the scale factor between SSIM~\cite{wang2004image} and L1 loss.
As described in~\cite{wang2004image}, the SSIM loss can measure structure similarity of images regardless of a complex illumination change.
We empirically found that the SSIM loss is also effective for thermal images because measured temperature values could temporally vary depending on the distances and heat sources.

{\bf Invalid Pixel Masking.}
This pixel-wise reconstruction signal is filtered out according to the following equation,~\eqnref{equ:recon mask loss}.
\begin{equation} 
\label{equ:recon mask loss}
L_{rec}^{T} = \frac{1}{|V_{p}|}\sum_{p\in V_{p}}{M^{T}_{gc}(p)\cdot M^{T}_{ns}(p)\cdot L_{rec}^{T}(p)},
\end{equation}
where $V_{p}$ stands for valid points that are successfully projected from $I_{t+1}^{T}$ to the image plane of $I_{t}^{T}$, $|V_{p}|$ defines the number of points in $V_{p}$, and $M_{gc}$ is the geometrically consistent pixel mask~\cite{bian2019unsupervised}, defined as $M_{gc}=1-D_{diff}$, used to exclude moving objects and occluded regions that may impair network training. The depth difference $D_{diff}$ will be described in the section on geometric consistency loss $L_{gc}$. Lastly, the non-static pixel mask $M_{ns}$~\cite{godard2019digging}, defined as $M_{ns} = [L_1({{I}_{t},\Tilde{I}_{t}}) < L_1({{I}_{t},I_{t+1}})]$, excludes pixels that remain the same between adjacent frames because of static camera motion and low texture regions.

\subsection{Photometric Consistency Loss}
Even if the temperature consistency loss can provide some extent supervisory signal, this signal is relatively weak compared to the visible images.
Therefore, visible image reconstruction loss can satisfactorily complement the temperature consistency loss.
However, usual visible image synthesis methods~\cite{zhou2017unsupervised,xu2019quadratic} essentially require a depth map and pose based on the visible image coordinate system that needs another neural network. 
Therefore, we propose a new visible image synthesis method that utilizes a heterogeneous coordinate system's depth map and poses without additional network burden. 

{\bf Forward Depth and Pose Warping.}
Initially, we generate the forward rigid flow $F_{t}^{T \shortrightarrow{} V}$ by exploiting the rigid geometric relationship between the visible and thermal camera.
With the predicted depth map $D_{t}^{T}$ and extrinsic matrix $P^{T \shortrightarrow{} V}$, the forward rigid flow $F_{t}^{T \shortrightarrow{} V}$ can be derived from the projective geometry as follows:
\begin{equation} 
\label{equ:rigid flow}
F_{t}^{T \shortrightarrow{} V}(p_{t}) = K^{V}P^{T \shortrightarrow{} V}D_{t}^{T}(p)(K^{T})^{-1}p_{t} - p_{t} 
\end{equation}
where $p_{t}$ is the homogeneous coordinates of pixels in $D_{t}^{T}$ and $K^{V}$,$K^{T}$ are the camera intrinsic matrices.
After that, we convert the forward flow $F_{t}^{T \shortrightarrow{} V}$ to generate the pseudo backward rigid flow $\Tilde{F}_{t}^{V \shortrightarrow{} T}$, as follows:
\begin{equation} 
\label{equ:flow reversa}
\Tilde{F}^{V \shortrightarrow{} T} = f_{reverse}(F^{T \shortrightarrow{} V}),
\end{equation}
where the function $f_{reverse}(\cdot)$ is the flow reversal layer~\cite{xu2019quadratic} that can convert the optical flow in the opposite way. The layer is also differentiable and allows gradients to be back-propagated. 
Finally, the depth $\Tilde{D}_{t}^{V}$ and pose $P_{t \shortrightarrow{} t+1}^{V}$ in the visible image coordinate system are estimated as follows :
\begin{gather} 
\Tilde{D}_{t}^{V} = W(D_{t}^{T}, \Tilde{F}_{t}^{V \shortrightarrow{} T}), \label{equ:visible depth generation} \\
\begin{split}
P_{t \shortrightarrow{} t+1}^{V} = P_{t+1}^{T \shortrightarrow{} V}P_{t \shortrightarrow{} t+1}^{T}P_{t}^{V \shortrightarrow{} T}, \label{equ:visible pose generation} \\
\end{split}
\end{gather}
where $ \Tilde{F}_{t}^{V \shortrightarrow{} T}$ represents pseudo pixel correspondences between $D_{t}^{T}$ and $D_{t}^{V}$, $P_{t}^{V \shortrightarrow{} T}$ and $P_{t+1}^{T \shortrightarrow{} V}$ are the extrinsic matrix $P^{T \shortrightarrow{} V}$ in an inverse relation, and $W(a,b)$ is an inverse warping function~\cite{jaderberg2015spatial} that transfers $a$ with pixel offset $b$, then conducts bi-linear interpolation. 

{\bf Visible Image Reconstruction.}
Based on the pseudo depth $\Tilde{D}_{t}^{V}$ and pose $P_{t \shortrightarrow{} t+1}^{V}$ in the visible image coordinate system, a synthesized visible image $\Tilde{I}_{t}^{V}$ is reconstructed from the source image $I_{t+1}^{V}$ in an inverse-warping manner~\cite{zhou2017unsupervised}. 
The photometric consistency loss can be obtained and filtered in the same manner as in~\eqnref{equ:recon loss} and ~\eqnref{equ:recon mask loss}.

\subsection{Geometric Consistency Loss}
The geometric consistency loss~\cite{bian2019unsupervised} enforces the consecutive depth maps $D_t$ and $D_{t+1}$ to conform the scale-consistent 3D scene structure.
The depth inconsistency map $D_{diff}$ is defined as:
\begin{equation} 
\label{equ:geometric consistency}
D_{diff}(p) = \frac{|\Tilde{D}_t(p)-D_t^{'}(p)|}{\Tilde{D}_t(p)+D_t^{'}(p)},
\end{equation}
where $\Tilde{D}_t(p)$ is the synthesized depth map generated from the source depth map $D_{t+1}$ and relative pose $P_{t \shortrightarrow{} t+1}$.
$D_t^{'}$ is the compensated depth map of $D_t$; 
since the depth value of the same object at time $t+1$ has already changed according to the motion $P_{t \shortrightarrow{} t+1}$, it needs to be compensated with the motion $P_{t \shortrightarrow{} t+1}$ at time $t$.
With the inconsistency map $D_{diff}$, the geometry consistency loss is simply defined as :
\begin{equation} 
\label{equ:geometric consistency loss}
L_{gc} = \frac{1}{|V_p|}\sum_{p\in V_p}{D_{diff}(p)},
\end{equation}
The geometric consistency loss $L_{gc}$ minimizes the geometric distance of the predicted depth between each consecutive pair and enforces their scale-consistency. During the training process, the consistency can propagate to the entire video sequence. 
The loss $L_{gc}$ is utilized for both estimated and synthesized depth maps from the thermal and visible image coordinate systems.



\section{Experimental Results}
\label{sec:result}
\begin{table*}[t]
\caption{\textbf{Quantitative comparison for depth estimation using ViViD dataset~\cite{alee-2019-icra-ws}}. 
We compare our network $Ours\text{-}T$ and $Ours\text{-}MS$ with state-of-the-art supervised/{self-supervised} depth network~\cite{ranftl2020towards,bian2019unsupervised}. 
Modality indicates the input source of each network. 
Supervision denotes supervision source, such as direct depth GT and self-supervision from each spectral image.
Overall, $Ours\text{-}MS$ outperforms both indoor and outdoor set by showing the lowest error results and the highest accuracy results.
}
\begin{center}
\resizebox{0.99 \linewidth}{!}{
\def\arraystretch{1.2}
\footnotesize
\begin{tabular}{|c|c|c|c|c|cccc|ccc|}
\hline
\multirow{2}{*}{Scene} & \multirow{2}{*}{Methods} & \multirow{2}{*}{Modality} & \multirow{2}{*}{Supervision} &\multirow{2}{*}{Cap}  & \multicolumn{4}{c|}{\textbf{Error $\downarrow$}} & 
\multicolumn{3}{c|}{\textbf{Accuracy $\uparrow$}}    \\ \cline{6-12}
 &  &  &  &  & AbsRel & SqRel & RMS & RMSlog &  $\delta < 1.25$ & $\delta < 1.25^{2}$ & $\delta < 1.25^{3}$ \\ 
\hline \hline
\multirow{9}{*}{\rotatebox{90}{\textbf{Indoor Well-lit}}}
    & {DispResNet (ResNet18)}   & RGB & Depth & {0-10m} & {0.263} & {0.502} & {0.515} & {0.259} & {0.847} & {0.958} & {0.985} \\
    & Midas-v2 (EfficientNet-Lite3)~\cite{ranftl2020towards}   & RGB & Depth & 0-10m & 0.198 & 0.355 & 0.383 & 0.216 & 0.919 & 0.979 & 0.991 \\
    & Midas-v2 (ResNext101)~\cite{ranftl2020towards} & RGB &  Depth & 0-10m & 0.194 & 0.348 & 0.370 & 0.210 & 0.928 & 0.979 & 0.991 \\
    & {DispResNet (ResNet18)}   & {Thermal} & Depth & {0-10m} & {0.117} & {0.097} & {0.462} & {0.170} & {0.869} & {0.960} & {0.991} \\
    & Midas-v2 (EfficientNet-Lite3)   & Thermal &  Depth & 0-10m & 0.062 & 0.044 & 0.282 & 0.107 & 0.946 & 0.983 & 0.995 \\
    & Midas-v2 (ResNext101) &  Thermal &  Depth & 0-10m & \textbf{0.057} & \textbf{0.039} & \textbf{0.269} & \textbf{0.102} & \textbf{0.954} & \textbf{0.984} & \textbf{0.995} \\
    \cline{2-12}
    & Bian~\etal(ResNet18)~\cite{bian2019unsupervised} & RGB & RGB & 0-10m & 0.327  & 0.532 & 0.715 & 0.306 & 0.661 & 0.932 & 0.979 \\
    & $Ours\text{-}T$ (ResNet18) & Thermal & Thermal & 0-10m & 0.225 & 0.201 & 0.709 & 0.262 & 0.620 & 0.920 & 0.993 \\
    & $Ours\text{-}MS$ (ResNet18) & Thermal & RGB\&T & 0-10m & \textbf{0.156} & \textbf{0.111} & \textbf{0.527} & \textbf{0.197} & \textbf{0.783} & \textbf{0.975} & \textbf{0.997} \\
\hline
\hline
\multirow{9}{*}{\rotatebox{90}{\textbf{Indoor Dark}}}
    & {DispResNet (ResNet18)}   & {RGB} & Depth & {0-10m} & {0.351} & {0.580} & {0.784} & {0.326} & {0.608} & {0.890} & {0.975} \\
    & Midas-v2 (EfficientNet-Lite3)~\cite{ranftl2020towards}   & RGB & Depth & 0-10m & 0.343 & 0.528 & 0.763 & 0.321 & 0.610 & 0.894 & 0.979 \\
    & Midas-v2 (ResNext101)~\cite{ranftl2020towards} & RGB &  Depth & 0-10m & 0.351 & 0.545 & 0.766 & 0.327 & 0.624 & 0.875 & 0.976\\
    & {DispResNet (ResNet18)}   & {Thermal} & Depth & {0-10m} & {0.124} & {0.094} & {0.466} & {0.174} & {0.854} & {0.963} & {0.992} \\
    & Midas-v2 (EfficientNet-Lite3)   & Thermal &  Depth & 0-10m & 0.060 & 0.036 & 0.273 & 0.105 & 0.950 & 0.985 & 0.996 \\
    & Midas-v2 (ResNext101) &  Thermal &  Depth & 0-10m & \textbf{0.053} & \textbf{0.032} & \textbf{0.257} & \textbf{0.099} & \textbf{0.958} & \textbf{0.987} & \textbf{0.996} \\
    \cline{2-12}
    & Bian~\etal(ResNet18)~\cite{bian2019unsupervised} & RGB & RGB & 0-10m & 0.452 & 0.803 & 0.979 & 0.399 & 0.493 & 0.786 & 0.933 \\
    & $Ours\text{-}T$ (ResNet18) & Thermal & Thermal & 0-10m & 0.232 & 0.222 & 0.740 & 0.268 & 0.618 & 0.907 & 0.987 \\
    & $Ours\text{-}MS$ (ResNet18) & Thermal & RGB\&T & 0-10m & \textbf{0.166} & \textbf{0.129} & \textbf{0.566} & \textbf{0.207} & \textbf{0.768} & \textbf{0.967} & \textbf{0.994} \\
\hline
\hline
\multirow{9}{*}{\rotatebox{90}{\textbf{Outdoor Night}}}
    & {DispResNet (ResNet18)}   & {RGB} & Depth & {0-80m} & {0.365} & {3.926} & {8.849} & {0.386} & {0.472} & {0.729} & {0.886} \\
    & Midas-v2 (EfficientNet-Lite3)~\cite{ranftl2020towards}   & RGB     &  Depth & 0-80m & 0.278 & 2.382 & 7.203 & 0.318 & 0.560 & 0.821 & 0.946 \\
    & Midas-v2 (ResNext101)~\cite{ranftl2020towards} & RGB     &  Depth & 0-80m & 0.264 & 2.187 & 7.110 & 0.306 & 0.571 & 0.833 & 0.955 \\
    & {DispResNet (ResNet18)}   & {Thermal} & Depth & {0-80m} & {0.159} & {1.101} & {5.019} & {0.212} & {0.857} & {0.964} & {0.980} \\
    & Midas-v2 (EfficientNet-Lite3)   & Thermal &  Depth & 0-80m & 0.090 & 0.464 & 3.385 & 0.130 & 0.910 & 0.981 & 0.995 \\
    & Midas-v2 (ResNext101) & Thermal &  Depth & 0-80m & \textbf{0.078} & \textbf{0.369} & \textbf{3.014} & \textbf{0.118} & \textbf{0.933} & \textbf{0.988} & \textbf{0.996} \\
    \cline{2-12}
    & Bian~\etal(ResNet18)~\cite{bian2019unsupervised} & RGB & RGB & 0-80m & 0.617 & 9.971    & 12.000 & 0.595 & 0.400 & 0.587 & 0.720 \\
    & $Ours\text{-}T$ (ResNet18) & Thermal & Thermal & 0-80m & 0.157	& 1.179	& 5.802	& 0.211	& 0.750	& 0.948	& 0.985 \\
    & $Ours\text{-}MS$ (ResNet18) & Thermal & RGB\&T & 0-80m & \textbf{0.146}	&  \textbf{0.873}	&  \textbf{4.697}	&  \textbf{0.184}	&  \textbf{0.801}	&  \textbf{0.973}	&  \textbf{0.993} \\
\hline
\end{tabular}
}
\end{center}
\label{tab:depth result}
\end{table*}

\begin{figure*}[t]
\begin{center}
    \resizebox{0.99 \linewidth}{!}{
        \footnotesize
        \begin{tabular}{c@{\hskip 0.004\linewidth}c@{\hskip 0.004\linewidth}c@{\hskip 0.004\linewidth}c@{\hskip 0.004\linewidth}c@{\hskip 0.004\linewidth}c@{\hskip 0.004\linewidth}c@{\hskip 0.004\linewidth}}
         \includegraphics[width=0.15\linewidth]{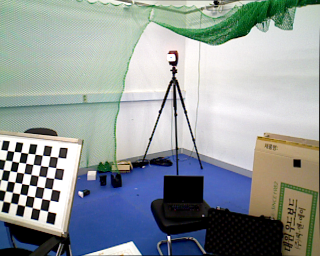} &
         \includegraphics[width=0.15\linewidth]{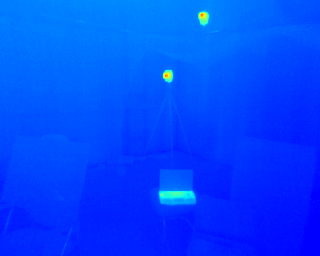} &
         \includegraphics[width=0.15\linewidth]{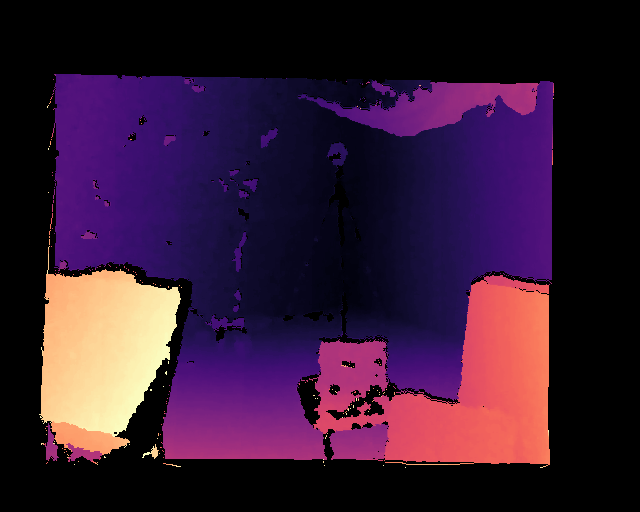} &
         \includegraphics[width=0.15\linewidth,height=2.12cm]{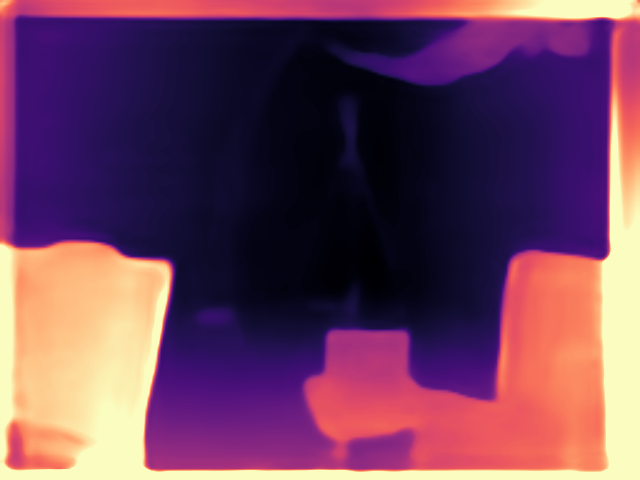} &
         \includegraphics[width=0.15\linewidth]{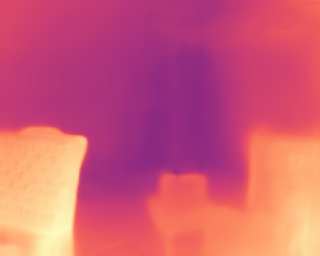} &
         \includegraphics[width=0.15\linewidth]{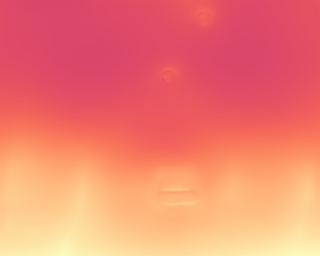} &
         \includegraphics[width=0.15\linewidth]{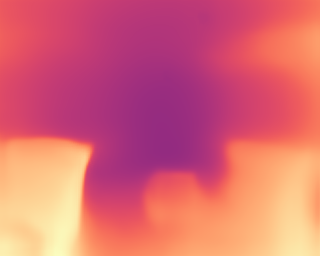} \vspace{-0.03in} \\
         \includegraphics[width=0.15\linewidth]{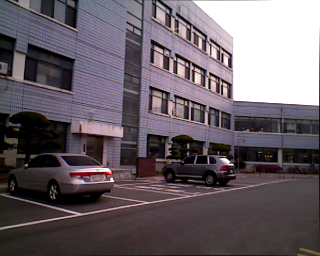} &
         \includegraphics[width=0.15\linewidth]{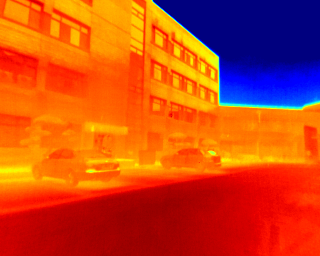} &
         \includegraphics[width=0.15\linewidth]{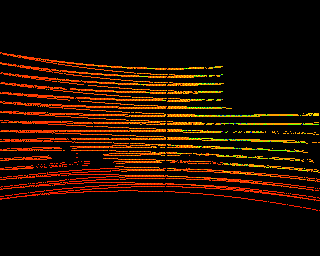} &
         \includegraphics[width=0.15\linewidth,height=2.12cm]{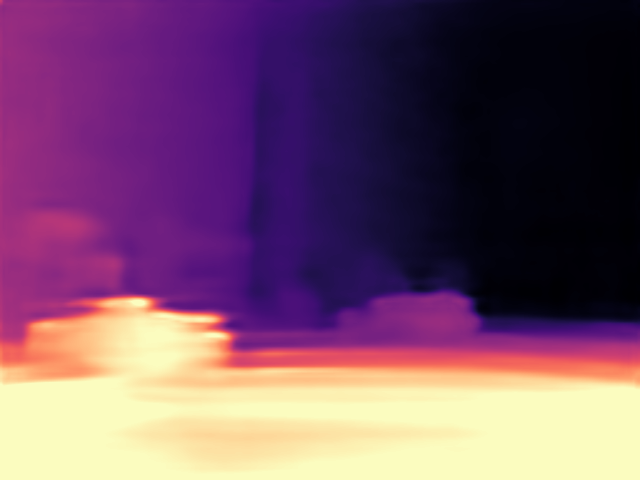} &
         \includegraphics[width=0.15\linewidth]{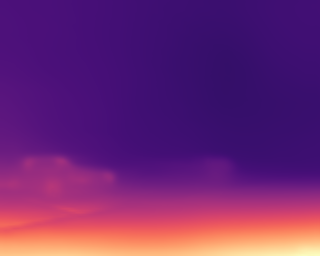} &
         \includegraphics[width=0.15\linewidth]{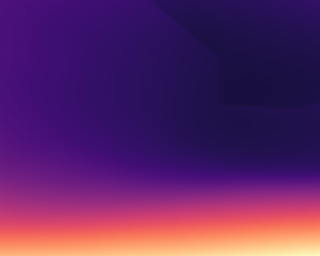} &
         \includegraphics[width=0.15\linewidth]{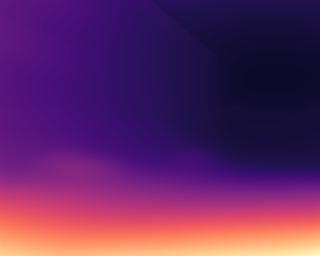} \vspace{-0.03in}  \\
         \includegraphics[width=0.15\linewidth]{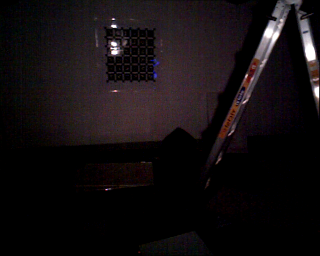} &
         \includegraphics[width=0.15\linewidth]{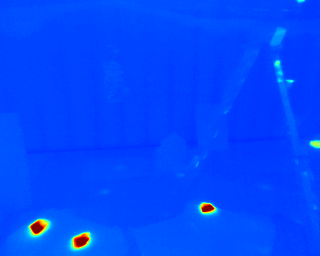} &
         \includegraphics[width=0.15\linewidth]{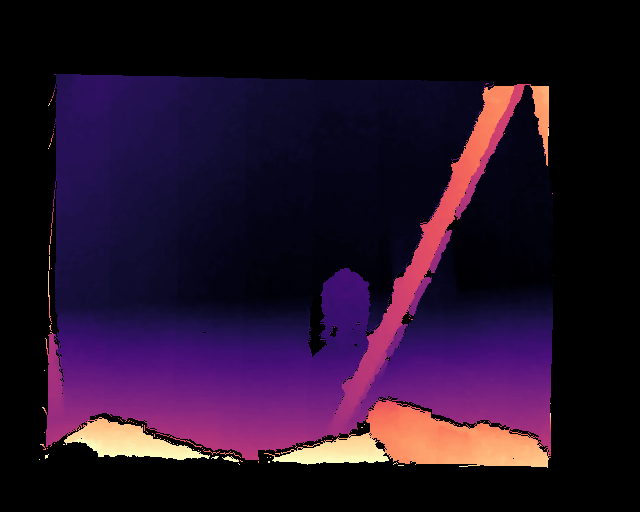} &
         \includegraphics[width=0.15\linewidth,height=2.12cm]{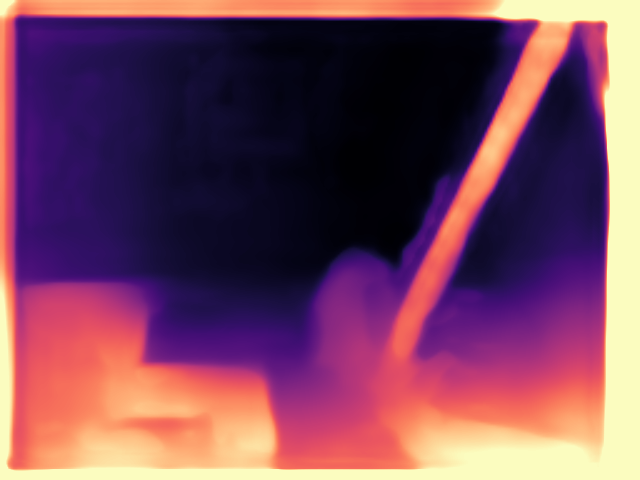} &
         \includegraphics[width=0.15\linewidth]{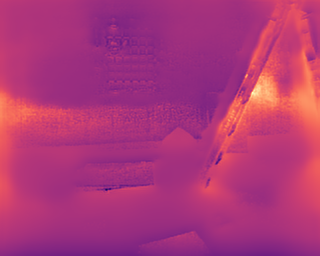} &
         \includegraphics[width=0.15\linewidth]{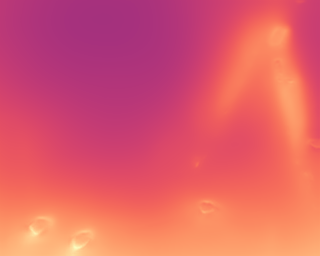} &
         \includegraphics[width=0.15\linewidth]{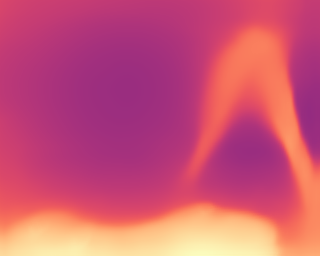} \vspace{-0.03in} \\
         \includegraphics[width=0.15\linewidth]{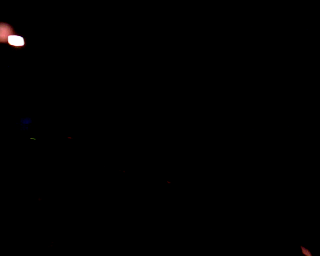} &
         \includegraphics[width=0.15\linewidth]{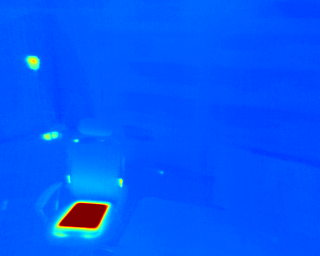} &
         \includegraphics[width=0.15\linewidth]{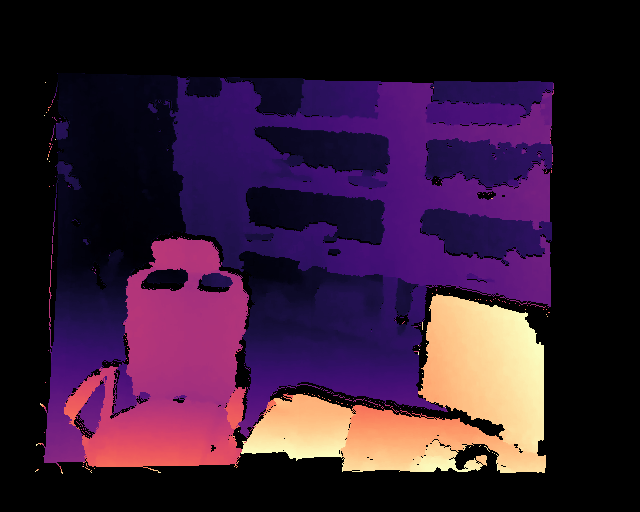} &
         \includegraphics[width=0.15\linewidth,height=2.12cm]{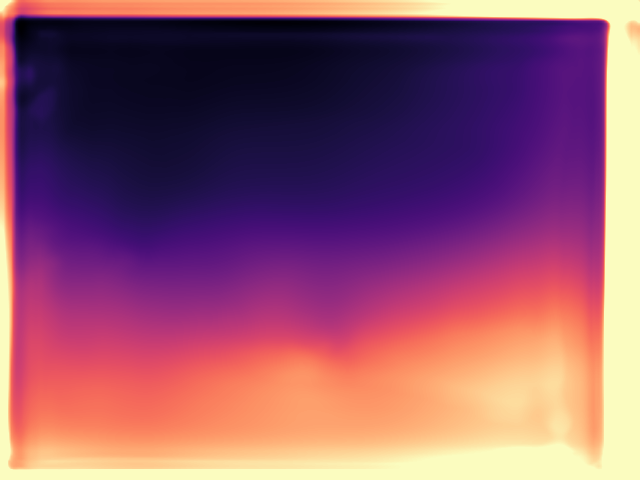} &
         \includegraphics[width=0.15\linewidth]{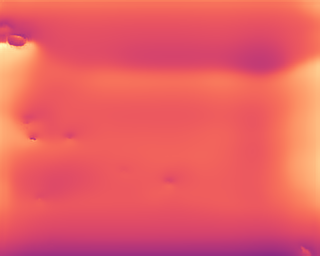} &
         \includegraphics[width=0.15\linewidth]{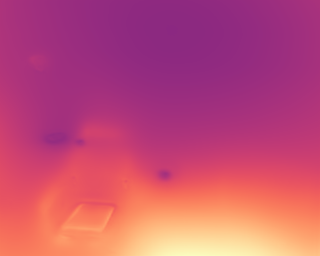} &
         \includegraphics[width=0.15\linewidth]{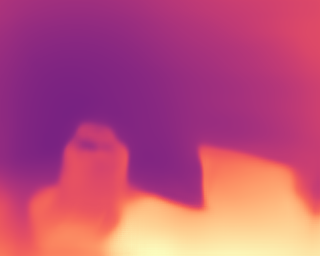} \vspace{-0.03in} \\
         \includegraphics[width=0.15\linewidth]{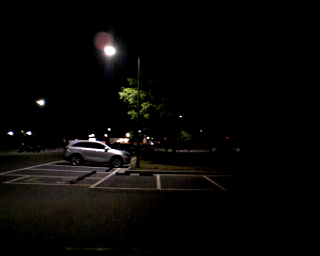} &
         \includegraphics[width=0.15\linewidth]{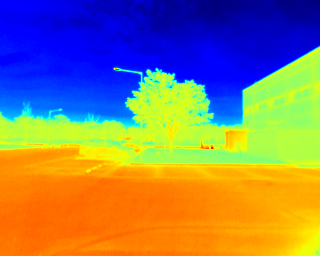} &
         \includegraphics[width=0.15\linewidth]{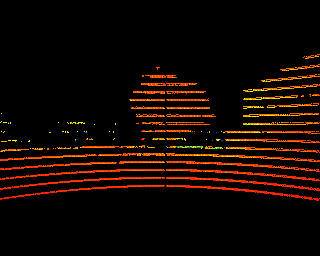} &
         \includegraphics[width=0.15\linewidth,height=2.12cm]{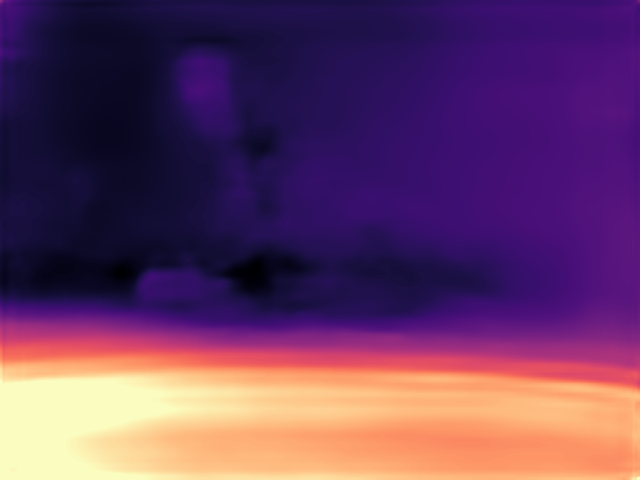} &
         \includegraphics[width=0.15\linewidth]{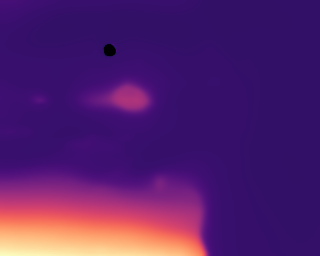} &
         \includegraphics[width=0.15\linewidth]{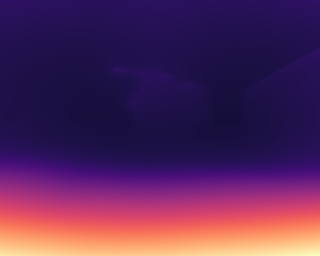} &
         \includegraphics[width=0.15\linewidth]{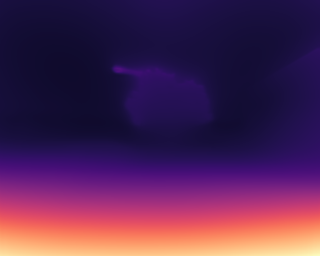} \vspace{-0.03in} \\
         \includegraphics[width=0.15\linewidth]{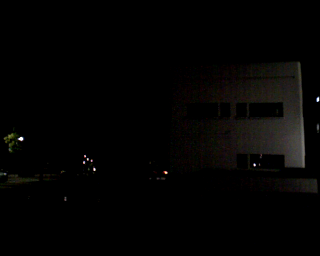} &
         \includegraphics[width=0.15\linewidth]{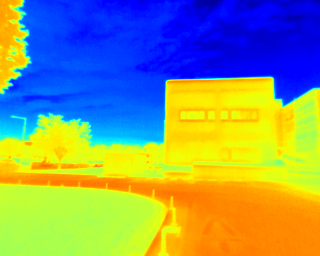} &
         \includegraphics[width=0.15\linewidth]{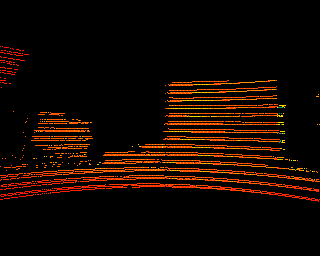} &
         \includegraphics[width=0.15\linewidth,height=2.12cm]{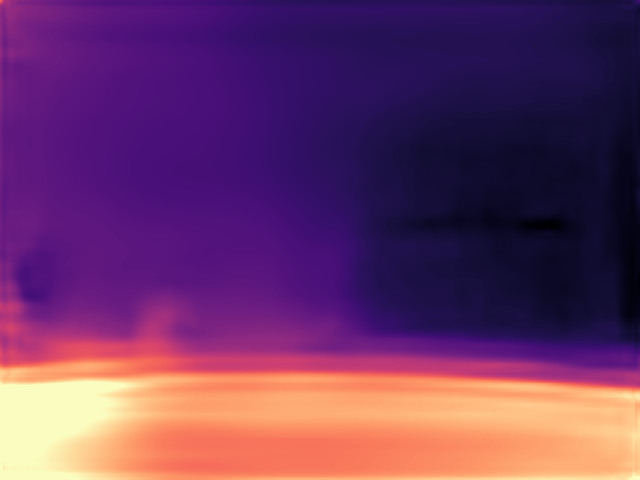} &
         \includegraphics[width=0.15\linewidth]{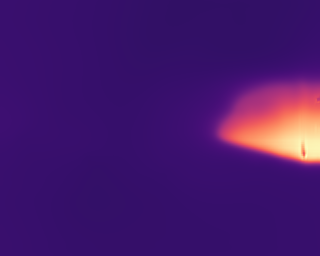} &
         \includegraphics[width=0.15\linewidth]{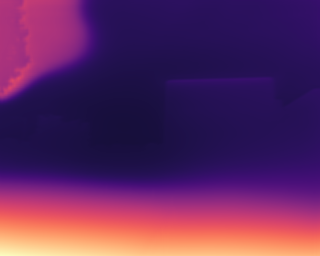} &
         \includegraphics[width=0.15\linewidth]{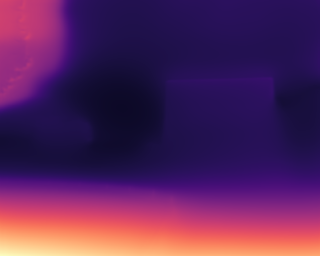} \vspace{-0.03in} \\
         {\footnotesize(a) Visible Image} & {\footnotesize (b) Thermal Image } & {\footnotesize (c) Ground Truth} & {\footnotesize (d) Midas v2} & {\footnotesize (e) Bian~\etal~\cite{bian2019unsupervised}} & {\footnotesize (f) $Ours\text{-}T$} & {\footnotesize (g) $Ours\text{-}MS$}\\
        \end{tabular}
    }
\end{center}
\caption{{\bf Qualitative comparison of depth estimation results on ViViD dataset~\cite{alee-2019-icra-ws}}. From left to right: visible images, thermal images, ground-truth depths, and depth map results obtained with Midas-v2 (RGB)~\cite{ranftl2020towards}, Bian~\etal~\cite{bian2019unsupervised}, $Ours\text{-}T$, and $Ours\text{-}MS$. 
Midas-v2~\cite{ranftl2020towards} and Bian~\etal~\cite{bian2019unsupervised} use visible images, and $Ours\text{-}T$ and $Ours\text{-}MS$ use thermal images as input.
The first two rows are the training set results taken under well-lit conditions. 
The other rows are test set results taken under low-light conditions. 
The results show that $Ours\text{-}T$ robustly estimates the depth regardless of the light conditions. 
Also, $Ours\text{-}MS$ provides accurate and sharp depth results thanks to the proposed multi-spectral consistency loss.} 
\label{fig:Exp_depth}
\end{figure*}


\subsection{Implementation Details}
{\bf ViViD dataset~\cite{alee-2019-icra-ws}.}
In order to train depth and pose networks based on the proposed learning method, it is essential to have a proper dataset that contains raw thermal images, RGB images, pose Ground-Truth(GT) labels, depth GT labels, and calibration results.
The satisfactory dataset, among the recently proposed datasets~\cite{treible2017cats, choi2018kaist, bijelic2020seeing, khattak2020keyframe,  kim2018multispectral, alee-2019-icra-ws}, is the ViViD dataset~\cite{alee-2019-icra-ws}, which provides multi-modal sensor data streams, including a thermal camera, an RGB-D camera, an event camera, an IMU, VICON, and a Lidar. 
The dataset consists of 10 indoor and 4 outdoor sequences with different illumination and motion conditions; the indoor set consists of slow, aggressive, and unstable motion with global, local, varying, and dark illumination, the outdoor set consists of only slow motion with day and night conditions.

\begin{table}[t]
\caption{\textbf{Training/Test Set Configuration of ViViD dataset.} 
We divide ViViD dataset with illumination conditions; the training set includes global, local, and day-time conditions. The test set includes local, dark, and night-time conditions.
}
\begin{center}
\resizebox{0.99\linewidth}{!}
{
    \def\arraystretch{1.3}
    \footnotesize
    \begin{tabular}{|c|c|c|c|c|}
    \hline
    \multicolumn{2}{|c|}{Scene} &  Motion & Illumination & $\#$ of images \\ 
    \hline
    \hline
    \multirow{3}{*}{Indoor}
    & Training set          & Slow, Unstable & Global, Local & 2178 \\
    & Test set (well-lit)   & Aggressive, Unstable & Local & 478 \\
    & Test set (dark)       & Slow,Aggressive, Unstable  & Dark & 1201 \\
    \hline
    \multirow{2}{*}{Outdoor}
    & Training set        & Slow & Day & 2213 \\
    & Test set (night)    & Slow & Night & 2019 \\
    \hline
    \end{tabular}
}
\end{center}
\label{tab:dataset}
\end{table}

{\bf Dataset generation.}
However, the dataset is provided in the form of a ROSbag. 
We conduct pre-processing to generate training and testing datasets.
We generate the indoor and outdoor depth GT labels in the thermal image coordinate system by transforming RGB-D and Lidar sensor data with extrinsic parameters.
The indoor and outdoor pose GT are transformed from VICON and Lidar SLAM~\cite{shan2018lego} results.
The training/test dataset configurations are shown in~\tabref{tab:dataset}.

{\bf Network architecture.}
We adopt DispResNet~\cite{ranjan2019competitive} and PoseNet~\cite{ranjan2019competitive} with the ResNet-18 encoder~\cite{he2016deep} to train single-view depth network and multi-view pose network.
We modify the first convolution layers of the original DispResNet and PoseNet to take single-channel input of thermal images.
The depth network takes a single monocular thermal image as input and predicts a depth map as output.
The pose network estimates a 6D relative camera pose from consecutive thermal images.
The networks and proposed learning method are all implemented with the PyTorch library~\cite{paszke2017automatic}. 

{\bf Training setup.}
We trained the depth and pose networks for 150 epochs on the single RTX titan GPU with 24GB memory. 
It took about 24 hours to train the networks.
Throughout the whole set of experiments, we set the hyperparameters $\alpha$ to $1.0$ and $\beta$ to $0.5$.
We used $\gamma^{T}$, $\gamma^{V}$, $\lambda_{T}$, $\lambda_{V}$, $\tau_{min}$, $\tau_{max}$ values of [0.15, 0.85, 0.25, 1.0, 10\textdegree C, 40\textdegree C] in the indoor set and [0.85, 0.30, 1.0, 0.1, 0\textdegree C, 30\textdegree C] in the outdoor set.

\subsection{Single-view Depth Estimation Results}
To validate the effectiveness of the proposed learning method for thermal images, we trained state-of-the-art supervised/self-supervised depth networks~\cite{ranftl2020towards,bian2019unsupervised} with an RGB input source on the ViViD Dataset~\cite{alee-2019-icra-ws}.
Moreover, we trained supervised depth networks, Midas-v2~\cite{ranftl2020towards} and our baseline depth network (\ie, DispResNet), that takes thermal images as input source to investigate the upper bound of thermal image based depth estimation performance.
Please note that, as mentioned earlier, the existing self-supervised depth network for thermal images~\cite{kim2018multispectral} requires optically aligned visible-thermal images and a visible stereo image pair; it is not compatible with the currently available dataset.
$Ours\text{-}T$ and $Ours\text{-}MS$ take thermal images as input source and are trained with thermal image losses $L_{rec}^{T}$, $L_{gc}^{T}$ and multi-spectral image losses $L_{rec}^{T,V}$, $L_{gc}^{T,V}$, respectively. 
We use Eigen~\etal~\cite{eigen2014depth}'s evaluation metrics to measure the performance of the depth estimation results. 
In the indoor and outdoor sets, we follow NYU~\cite{silberman2012indoor} and KITTI~\cite{geiger2013vision} evaluation settings.

The experimental results are shown in~\tabref{tab:depth result} and~\figref{fig:Exp_depth}.
As shown in~\figref{fig:Exp_depth}, Midas-v2~\cite{ranftl2020towards} and Bian~\etal\cite{bian2019unsupervised} provide precise depth estimation results when sufficient light is guaranteed.
However, the performance significantly decreases as the illumination condition become worse.
On the other hand, $Ours\text{-}T$ can robustly estimate the depth map regardless of the lighting condition, but suffers from a relatively inaccurate depth quality.
Especially, this phenomenon frequently occurs in the indoor set because the indoor thermal image has homogeneous temperature distribution and distinctively high-temperature objects, causing high error signals.
However, the proposed learning method can manage this phenomenon by providing a visible-spectral consistency signal.
By leveraging both visible and thermal spectral images' advantage, $Ours\text{-}MS$ shows the lowest error results and accurate prediction results regardless of the lighting condition .

\begin{table*}[t]
\caption{\textbf{Quantitative comparison of pose estimation results on the ViViD dataset~\cite{alee-2019-icra-ws}}.
We compare our networks with ORB-SLAM2~\cite{mur2017orb} and Bian~\etal~\cite{bian2019unsupervised}.
Please note that ORB-SLAM2 often failed to track visible and thermal image sequences. 
Therefore, we calculated the accuracy using the valid parts of the sequences that are successfully tracked.
On the other hand, the accuracy of other networks is calculated using the whole sequence.
(\textcolor{black}{\textbf {Black}} : Best, \textcolor{blue}{Blue} : Runner-up).
}
\begin{center}
\resizebox{0.90\linewidth}{!}
{
    \def\arraystretch{1.0}
    \footnotesize
    \begin{tabular}{|c|c|cc|cc|cc|}
    \hline
    \multirow{2}{*}{Scene} & \multirow{2}{*}{Methods} & \multicolumn{2}{c|}{$M_{slow}$ + $I_{dark}$} & \multicolumn{2}{c|}{$M_{slow}$ + $I_{vary}$} & \multicolumn{2}{c|}{$M_{aggressive}$ + $I_{local}$} \\ 
    \cline{3-8} & & ATE & RE & ATE & RE & ATE & RE \\ 
    \hline \hline
    \multirow{5}{*}{\rotatebox{90}{\textbf{Indoor}}}
        & ORB-SLAM (V)  & - & - & 0.0089$\pm$0.0085 & 0.0102$\pm$0.0064 & 0.0319$\pm$0.0098 & \textbf{0.006$\pm$0.015} \\
        & ORB-SLAM (T)    & 0.0091$\pm$0.0066 & \textbf{0.0072$\pm$0.0035} & 0.0090$\pm$0.0088 & \textbf{0.0068$\pm$0.0034} & - & - \\
        & Bian~\etal~\cite{bian2019unsupervised} & 0.0064$\pm$0.0036 & 0.0211$\pm$0.0178 & 0.0073$\pm$0.0065 & 0.0332$\pm$0.0566 & 0.0312$\pm$0.0245 & 0.0667$\pm$0.0602 \\ 
        & $Ours\text{-}T$ &  \textcolor{blue}{0.0063$\pm$0.0029} & 0.0092$\pm$0.0056 & \textcolor{blue}{0.0067$\pm$0.0066} & \textcolor{blue}{0.0095$\pm$0.0111} & \textbf{0.0225$\pm$0.0125} & 0.0671$\pm$0.055 \\
        & $Ours\text{-}MS$ & \textbf{0.0057$\pm$0.003} & \textcolor{blue}{0.0089$\pm$0.005} & \textbf{0.0058$\pm$0.0032} & 0.0102$\pm$0.0124 & \textcolor{blue}{0.0279$\pm$0.0166} & \textcolor{blue}{0.0507$\pm$0.035} \\
    \hline
    \hline
    \multirow{2}{*}{Scene} & \multirow{2}{*}{Methods} & \multicolumn{2}{c|}{$M_{slow}$ + $I_{night}$ (1)} & \multicolumn{2}{c|}{$M_{slow}$ + $I_{night}$ (2)} & & \\ 
    \cline{3-8} & & ATE & RE & ATE & RE & & \\ 
    \hline
    \hline
    \multirow{5}{*}{\rotatebox{90}{\textbf{Outdoor}}}
        & ORB-SLAM (V)  & 0.2375$\pm$0.1607 & 0.0286$\pm$0.0136 & 0.1824$\pm$0.1168 & 0.0302$\pm$0.0143 & & \\
        & ORB-SLAM (T)    & 0.1938$\pm$0.1380 & 0.0298$\pm$0.0150 & 0.1767$\pm$0.1094 & 0.0287$\pm$0.0126 & & \\
        & Bian~\etal~\cite{bian2019unsupervised} & 0.0708+-0.0394 & 0.0302+-0.0142 & 0.0668$\pm$0.0376 & 0.0276$\pm$0.0121 & & \\
        & $Ours\text{-}T$ &  \textcolor{blue}{0.0571$\pm$0.0339} & \textbf{0.028$\pm$0.0139} & \textbf{0.0534$\pm$0.029} & \textbf{0.0272$\pm$0.0121} & & \\
        & $Ours\text{-}MS$ &  \textbf{0.0562$\pm$0.031} & \textcolor{blue}{0.0287$\pm$0.0144} & \textcolor{blue}{0.0598$\pm$0.0316} & \textcolor{blue}{0.0274$\pm$0.0124} & & \\
    \hline
    \end{tabular}
}
\end{center}
\label{tab:ATE_result}
\end{table*}

\subsection{Pose Estimation Results}
We compare our pose estimation networks $Ours\text{-}T$ and $Ours\text{-}MS$ with ORB-SLAM2~\cite{mur2017orb} and Bian~\etal~\cite{bian2019unsupervised} on the ViViD dataset~\cite{alee-2019-icra-ws}.
We consider two types of ORB-SLAM2~\cite{mur2017orb} that take visible or thermal image input, ORB-SLAM(V) and ORB-SLAM(T).
We utilize the 5-frame pose evaluation method~\cite{zhou2017unsupervised}. 
The evaluation metrics are Absolute Trajectory Error (ATE) and Relative Error (RE)~\cite{zhang2018tutorial}.
Since each sequence of the test dataset has different illumination and motion conditions, we evaluate the pose estimation performance on each sequence to investigate condition-wise performance differences.

The experimental results are shown in~\tabref{tab:ATE_result}. 
ORB-SLAM2 often failed to track visible and thermal image sequences, so we calculated the accuracy using the valid parts of the sequences that ORB-SLAM2 successfully tracked.
Overall, ORB-SLAM(V) and Bian~\etal~\cite{bian2019unsupervised} show comparable results when some extent of illumination condition is guaranteed, such as $I_{local}$. 
However, the pose estimation performance is degraded in the low-light conditions $I_{dark}$ and $I_{night}$.
The methods $Ours\text{-}T, Ours\text{-}MS$, and ORB-SLAM(T), using thermal image input, show consistent performance regardless of the lighting condition.
ORB-SLAM(T) provides accurate RE error in the indoor set because the high-temperature objects have distinctive features, while the performance is degraded in the outdoor set.
Overall, our networks $Ours\text{-}T$ and $Ours\text{-}MS$ can provide accurate and reliable pose estimation results regardless of motion and lighting conditions.

\begin{figure}[t]
\footnotesize
\begin{tabular}{@{}c@{\hskip 0.005\linewidth}c@{\hskip 0.005\linewidth}c}
\includegraphics[width=0.32\linewidth]{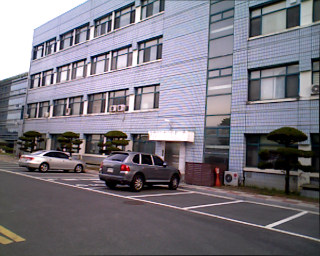} &
\includegraphics[width=0.32\linewidth]{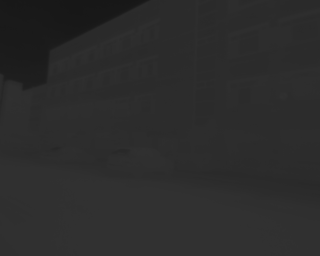} &
\includegraphics[width=0.32\linewidth]{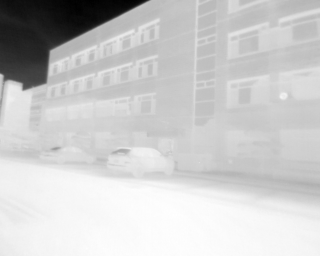} \\
(a) Visible image & (b) $N_{whole}$ & (c) $N_{MM}$ \vspace{0.05in}\\
\includegraphics[width=0.32\linewidth]{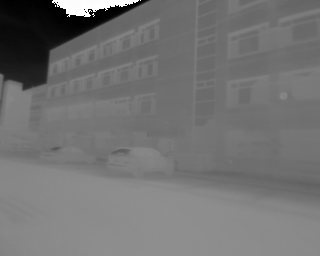} &
\includegraphics[width=0.32\linewidth]{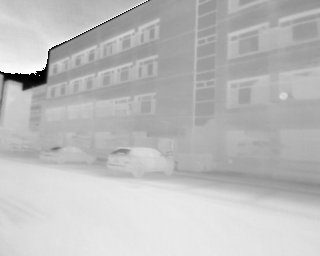} &
\includegraphics[width=0.32\linewidth]{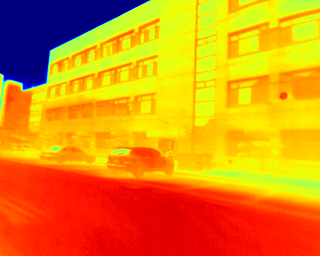} \\
(d) $N_{WC}$ & (e) $N_{NC}$ & (f) $N_{CC}$ \\
\end{tabular}
\caption{{\bf Illustration results according to thermal image normalization methods.} 
$N_{whole}$, $N_{MM}$, $N_{WC}$, $N_{NC}$, and $N_{CC}$ indicates normalization with the whole range, min-max range, widely clipped range, narrowly clipped range, and narrow clipped range and colorization, respectively.
}
\label{fig:abl_thr_normalize}
\end{figure}

\subsection{Ablation Study}
\label{sec:ablation}
{\bf Effects of thermal image representation methods.}
Unlike a visible image that has evenly distributed measurement values within its whole range, most measured values for the raw thermal image exist only in a specific and narrow range.
In this study, we investigate the effect of each thermal image representation method for self-supervised depth and pose learning from thermal images.
$N_{whole}$, $N_{MM}$, $N_{WC}$, $N_{NC}$, and $N_{CC}$ indicate that images are re-scaled in the whole range($2^{14}$),  min-max range, widely clipped range ($\tau_{min}$,$\tau_{max}$ = 0\textdegree C,60\textdegree C), narrowly clipped range, and narrow clipped range with colorization, respectively.
$N_{MM}$ is identical with the typical 8bit representation of thermal images, used in lots of datasets and recognition tasks~\cite{bijelic2020seeing, treible2017cats, choi2018kaist}.

The quantitative and qualitative results are shown in~\tabref{table:abl_Normalize_depth} and ~\figref{fig:abl_thr_normalize}.
$N_{whole}$ loses the whole image details leading to a weak self-supervisory signal, $N_{MM}$ has a higher contrast image than $N_{whole}$ but frequently violates the temporal consistency assumption leading to degraded depth performance, and $N_{WC}$ and $N_{NC}$ preserve the temporal temperature consistency and relatively high contrast image, which can provide a sufficient signal for network training.
However, the supervision signal of the continuous value in the single-channel is not sufficient. 
Therefore, we map a single-channel continuous value into a three-channel discontinuous value to generate a more stronger temperature consistency signal. 
Thanks to the $N_{CC}$, the depth estimation network shows better performance. 

\begin{table}[t]
 \centering
 \caption{\textbf{Depth results comparison for each thermal image normalization method.} Top to bottom: indoor and outdoor test set results of ViViD dataset~\cite{alee-2019-icra-ws}. 
 }
    \resizebox{0.98\linewidth}{!}
    {
        \def\arraystretch{1.1}
        \begin{tabular}{c|cccc|ccc}
        \hline
        \multirow{2}{*}{Methods} & \multicolumn{4}{c|}{\textbf{Error $\downarrow$}} & 
        \multicolumn{3}{c}{\textbf{Accuracy $\uparrow$}}    \\ \cline{2-8}
         & AbsRel & SqRel & RMS & RMSlog &  $< 1.25$ & $< 1.25^{2}$ & $< 1.25^{3}$ \\ 
        \hline \hline
        $N_{whole}$ & 0.255	& 0.292	& 0.838	& 0.285	& 0.602	& 0.884	& 0.980 \\
        $N_{MM}$ & 0.264	& 0.278	& 0.809	& 0.304	& 0.570	& 0.862	& 0.977 \\
        $N_{WC}$ & 0.259	& 0.301	& 0.853	& 0.317	& 0.598	& 0.879	& 0.975 \\ 
        $N_{NC}$ & 0.263	& 0.321	& 0.888	& 0.348	& 0.593	& 0.877	& 0.971 \\ 
        $N_{CC}$ & \textbf{0.231}	& \textbf{0.215}	& \textbf{0.730}	& \textbf{0.266}	& \textbf{0.616}	& \textbf{0.912}	& \textbf{0.990} \\
        \hline \hline
        $N_{whole}$ & 0.931	& 21.281 & 16.025	& 0.759	& 0.253	& 0.489	& 0.643 \\
        $N_{MM}$ & 0.552	& 7.875 & 10.616	& 0.539	& 0.440	& 0.633	& 0.767 \\
        $N_{WC}$ & 0.627	& 10.851 & 11.490	& 0.582	& 0.429	& 0.625	& 0.752 \\
        $N_{NC}$ & 0.158	& \textbf{1.178}	& \textbf{5.782}	& 0.211	& 0.748	& \textbf{0.950}	& 0.984 \\
        $N_{CC}$ & \textbf{0.157}	& 1.179	& 5.802	& 0.211	& \textbf{0.750}	& 0.948	& \textbf{0.985} \\
        \hline
        \end{tabular}
        \label{table:abl_Normalize_depth}
    }
\captionsetup{font=footnotesize}
\label{table:Abl_thr_normalize}
\end{table}

\begin{table}[t]
\centering
\caption{\textbf{Ablation study on loss functions of visible image ($L_{rec}^{V}$, $L_{gc}^{V}$, and $M^{V}$)}. 
Top to bottom: indoor and outdoor test set results on ViViD dataset~\cite{alee-2019-icra-ws}. 
}
\resizebox{0.98\linewidth}{!}
{
    \def\arraystretch{1.1}
    \begin{tabular}{c|ccc|c|ccc}
    \hline
    \multirow{2}{*}{Methods} & \multicolumn{3}{c|}{Loss functions} & \textbf{Error $\downarrow$} &  \multicolumn{3}{c}{\textbf{Accuracy $\uparrow$}} \\ \cline{2-8}  
     & $L_{rec}^{V}$ & $L_{gc}^{V}$ & $M^{V}$ & RMS & $< 1.25$ & $< 1.25^{2}$ & $< 1.25^{3}$ \\
    \hline
    \hline
    $Ours\text{-}T$  & & &                                & 0.730	& 0.616	& 0.912	& 0.990 \\
    {$+L_{rec}^{V}$}      &\checkmark &           &              & 0.614	& 0.702	& 0.955	& 0.993	\\
    {$+M^{V}$}      &\checkmark &           &\checkmark    & 0.554	& \textbf{0.788}	& 0.965	& 0.995\\
    $Ours\text{-}MS$ &\checkmark &\checkmark &\checkmark  &\textbf{0.553}	& 0.771	& \textbf{0.970}	& \textbf{0.995}\\
    \hline
    \hline
    $Ours\text{-}T$  & & &                                & 5.802	& 0.750	& 0.948	& 0.985 \\
    {$+L_{rec}^{V}$}      &\checkmark &           &              & 5.329	& 0.780	& 0.963	& 0.990 \\
    {$+M^{V}$}      &\checkmark &           &\checkmark    & 4.874	& 0.785	& 0.971	& 0.993 \\
    $Ours\text{-}MS$ &\checkmark &\checkmark &\checkmark  & \textbf{4.697}	& \textbf{0.801}	& \textbf{0.973}	& \textbf{0.993}	\\
    \hline
    \end{tabular}
}
\label{tab:abl_loss_functions}
\end{table}

{\bf Ablation study on loss functions.}
We conduct an ablation study about the loss functions $L_{rec}^{V}$, $L_{gc}^{V}$, and $M^{V}$, as shown in ~\tabref{tab:abl_loss_functions}.
By adding the forward-warping based photometric loss $L_{rec}^{V}$, the overall network performances both indoor and outdoor sets are improved.
Also, occlusions, static pixels, and homogeneous regions in the visible image are handled with the mask $M^{V}=M_{ns}^V\cdot M_{gc}^V$, leading to further performance improvement by minimizing the photometric loss in reliable regions.
Since the loss $L_{gc}^{V}$ is basically generated from the thermal image depth, the loss $L_{gc}^{V}$ gives a further depth consistency signal for thermal image's depth map. 
Based on the experimental results and RMS criteria, we select our final model $Ours\text{-}MS$, which utilizes $L_{rec}^{V}$, $L_{gc}^{V}$, and $M^{V}$.

\section{Conclusion}
\label{sec:conclusion}
In this paper, we propose a self-supervised learning method for single-view depth and multi-view pose estimation from a monocular thermal video.
The proposed learning method exploits temperature and photometric consistency loss to generate a self-supervisory signal.
The clipping-and-colorization strategy generates a sufficient self-supervisory signal from a thermal image, while preserving the temporal consistency.
To transfer complementary knowledge, the photometric consistency loss can synthesize a visible image with a depth map and pose estimated from a heterogeneous coordinate system.
Networks trained with the proposed method robustly estimate reliable and accurate depth and pose from monocular thermal video regardless of lighting conditions.
Our source code and the post-processed dataset we used are available at \url{https://github.com/UkcheolShin/ThermalSfMLearner-MS}.

\ifCLASSOPTIONcaptionsoff
  \newpage
\fi

{
\bibliographystyle{IEEEtran}
\bibliography{egbib}
}

\end{document}